\documentclass[11pt,a4paper]{article}
\usepackage[hyperref]{computel3}
\usepackage{times}
\usepackage{latexsym}

\usepackage{url}

\usepackage{amsmath}	% for \begin{align}
\usepackage{graphicx}	% for \includegraphics{filename}

\aclfinalcopy % Uncomment this line for the final submission

\title{Towards a General-Purpose Linguistic Annotation Backend}

\author{Graham Neubig$^\dag$, Patrick Littell$^\ddag$, Chian-Yu Chen$^\dag$, Jean Lee$^\dag$, \\ \textbf{Zirui Li$^\dag$, Yu-Hsiang Lin$^\dag$, Yuyan Zhang$^\dag$} \\
  $^\dag$Language Technologies Institute, Carnegie Mellon University \\
  $^\ddag$National Research Council Canada \\
  {\tt gneubig@cs.cmu.edu}}

\date{}

\begin{document}
\maketitle
% \begin{abstract}
% TODO
% \end{abstract}

\section{Introduction}

Language documentation is inherently a time-intensive process; transcription, glossing, and corpus management consume a significant portion of documentary linguists' work.  Advances in natural language processing can help to accelerate this work, using the linguists' past decisions as training material, but questions remain about how to prioritize human involvement.

In this extended abstract, we describe the beginnings of a new project that will attempt to ease this language documentation process through the use of natural language processing (NLP) technology.
It is based on (1) methods to adapt NLP tools to new languages, based on recent advances in massively multilingual neural networks, and (2) backend APIs and interfaces that allow linguists to upload their data (\S\ref{sec:framework}).
We then describe our current progress on two fronts: automatic phoneme transcription, and glossing (\S\ref{sec:progress}).
Finally, we briefly describe our future directions (\S\ref{sec:future}).

\section{Overall Framework}
\label{sec:framework}

The final goal of our project is to create a linguistic annotation backend (LAB), that will take in raw or partially annotated linguistic data, and provide annotation candidates for a linguist (or other interested party) to peruse.
Candidates for the types of services to provide are automatic phoneme transcription \cite{adams18lrec,michaud18ldc}, speech-to-text alignmen \cite{johnson2018forced}, word segmentation \cite{peng04chinesecrf,goldwater09bayesianws}, morphological analysis \cite{yarowsky00minimally}, syntactic analysis \cite{nivre2005dependency}, automatic glossing \cite{riding2008statistical}, or linguistic typology prediction \cite{daume07typology}.
The LAB will be hosted on a server and exposed as an API that can be linked to popular annotation software such as ELAN\footnote{\url{https://tla.mpi.nl/tools/tla-tools/elan/}} or FLEx.\footnote{\url{https://software.sil.org/fieldworks/}}

The obvious difficulty in creating such an interface is data scarcity in the languages in question.
In order to overcome these barriers, we plan to take advantage of recent advances in NLP that allow for multilingual modeling \cite{tackstrom12transfer,johnson16multilingual} and multi-task learning \cite{caruana1997multitask}, which allow models to be trained with very little, or even no data in the target language \cite{neubig18emnlp}.
We also plan to utilize active learning \cite{settles09alsurvey}, which specifically asks the linguists to focus on particular examples to maximize the effect of linguists' limited time when working with field data.
While there is still no alternative to significant human engagement when processing data, many of the decisions a linguist is faced with when transcribing, glossing, organizing, or searching a corpus are relatively rote -- decisions that could be deducible from past decisions or from similar languages.  

\section{Current Progress: A Backend/Interface for Automatic Phoneme Transcription and Glossing}
\label{sec:progress}

% WRITE (Amber/Zirui): Could you write one paragraph about the workflow and automatic interface? I'd also like to mention a little bit about what Oliver and Jannis are doing with regards to ASR, but I can add this later.

As first steps towards realizing our final goal, we have currently developed a backend for two tasks (automatic phoneme transcription and glossing), which is integrated with a simple example interface.

\begin{figure}[t]
\centering
\includegraphics[width=1.0\linewidth]{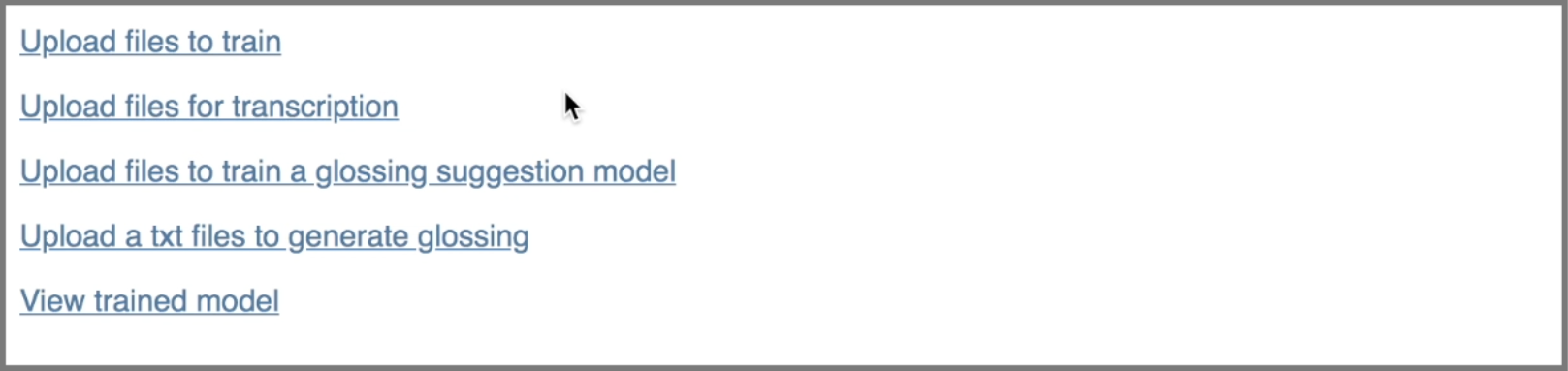}
\caption{The prototype of the automated interface supporting transcription and glossing.}
\label{fig:figInterface}
\end{figure}

\subsection{Backend Overview}
The current LAB is based on a simple three-step process:
\begin{description}
\item[Data Upload] The linguist uploads any existing annotated data to the interface.
\item[Model Training] A model is trained to process this data. This training could potentially utilize other data sources for multilingual and multi-task training.
\item[Data Annotation] The linguist uploads unannotated data, and the trained model proposes annotations for the linguist to accept or edit.
\end{description}
An example of an overall interface exposing this functionality for the currently implemented tasks of transcription and glossing is shown in Figure \ref{fig:figInterface}.

\subsection{Phoneme Recognition}
\begin{figure}[t]
\centering
\includegraphics[width=1.0\linewidth]{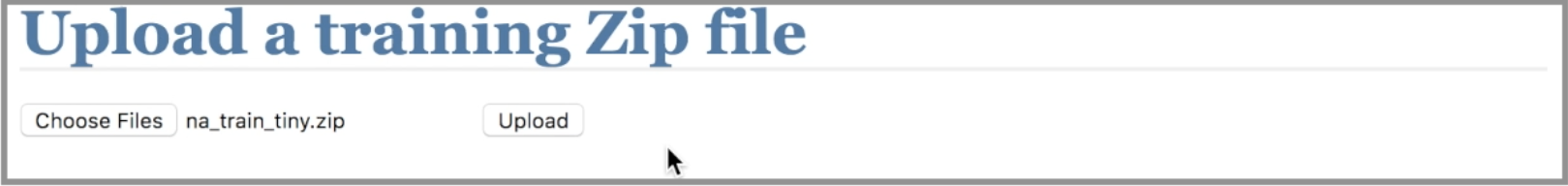}
\caption{Uploading the training data to the automated interface for transcription model training.}
\label{fig:figTranscribe1}
\end{figure}

\begin{figure}[t]
\centering
\includegraphics[width=1.0\linewidth]{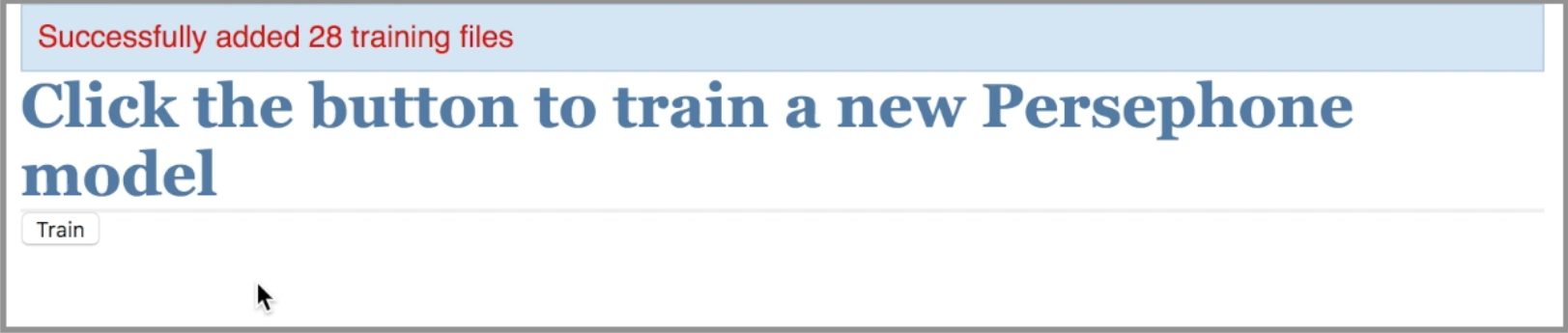}
\caption{Training a transcription model using the training data uploaded by the user.}
\label{fig:figTranscribe2}
\end{figure}

\begin{figure}[t]
\centering
\includegraphics[width=1.0\linewidth]{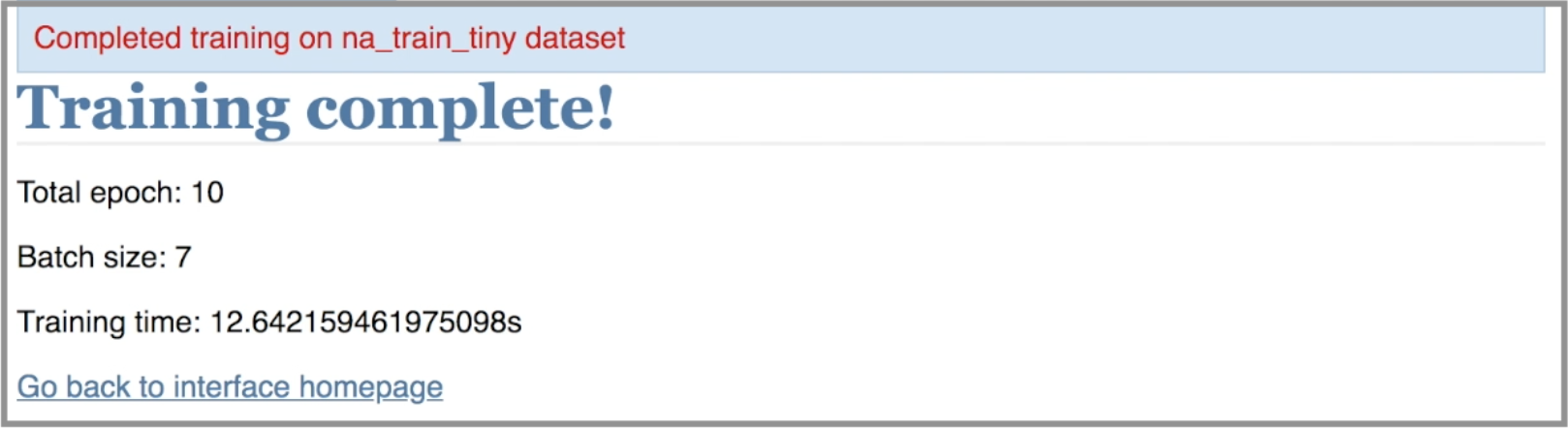}
\caption{Training a transcription model using the training data uploaded by the user.}
\label{fig:figTranscribe3}
\end{figure}

The automatic phoneme transcription component provides an interactive online interface for users to manage speech recognition models and transcribe speech recordings.
The speech recognition model can be any one of the user's choosing as long as it supports the API.
In our current system, we use Persephone \cite{adams18lrec} as our transcription backend, which is designed for low-resource language transcription.
Through the API, the users can upload a batch of speech recordings along with the corresponding transcriptions as the training data to train a transcription model tailored to the language and speakers of their interest.
The system is equipped with some default model and training configurations so that the users are not required to have expert knowledge of the transcription model and training.
The model obtained from each training session will then be stored for later use.
Figures \ref{fig:figTranscribe1} to \ref{fig:figTranscribe3} show the work flow of training a transcription model.

\begin{figure}[t]
\centering
\includegraphics[width=1.0\linewidth]{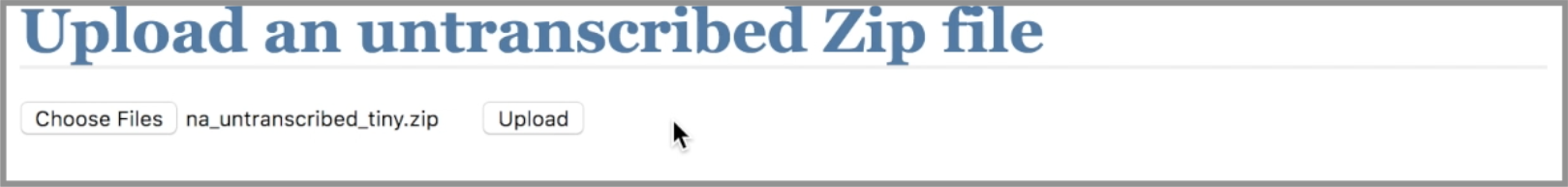}
\caption{Uploading the speech recordings to transcribe.}
\label{fig:figTranscribe4}
\end{figure}

\begin{figure}[t]
\centering
\includegraphics[width=1.0\linewidth]{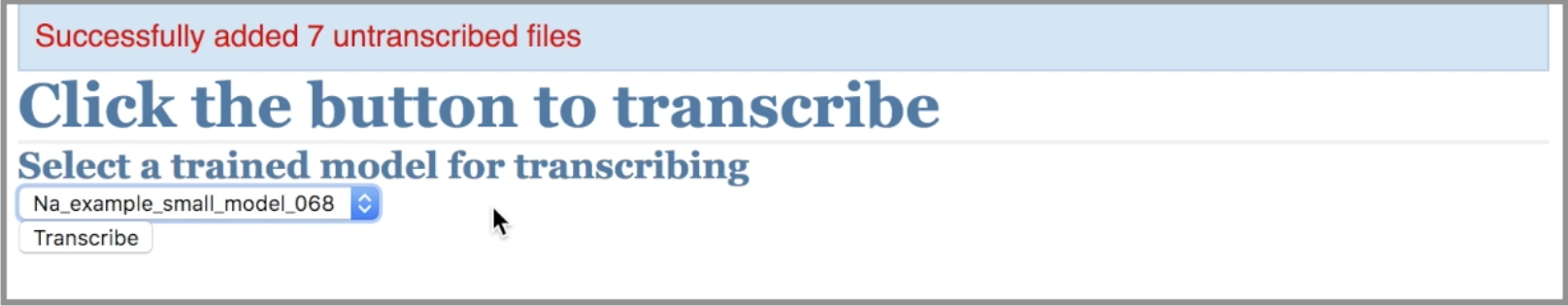}
\caption{Transcribing the speech recordings using the model previously trained.}
\label{fig:figTranscribe5}
\end{figure}

\begin{figure*}[t]
\centering
\includegraphics[width=1.0\linewidth]{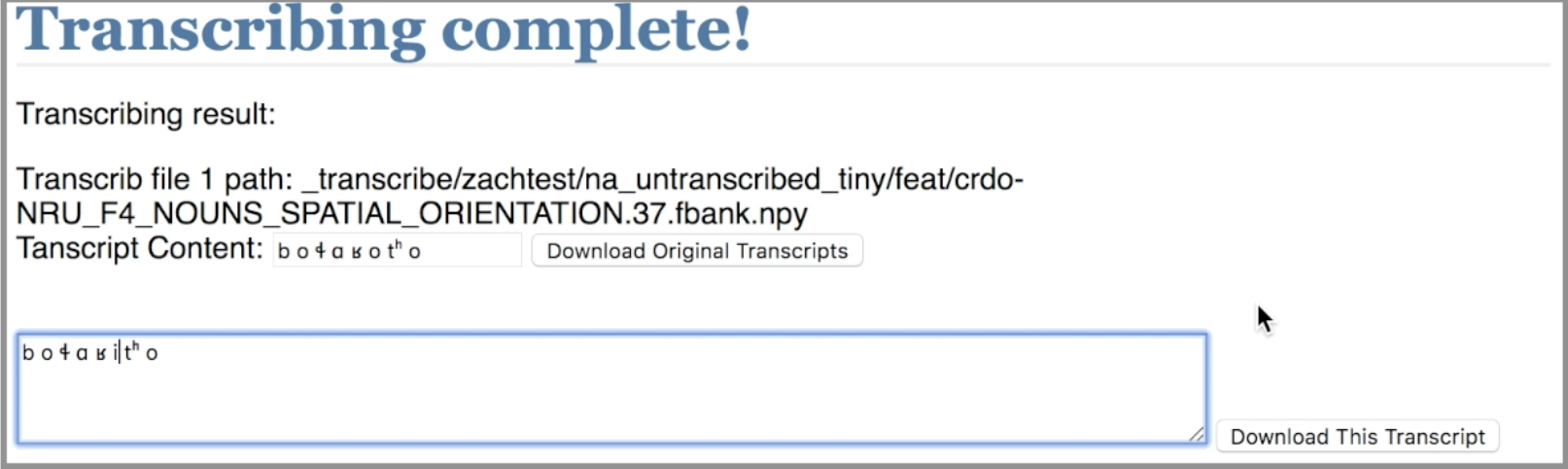}
\caption{The users can examine the transcription results and optionally edit the results to correct errors.}
\label{fig:figTranscribe6}
\end{figure*}

The users can upload speech recordings they want to transcribe to the interface, and perform the automatic transcription using previously trained models (Figures \ref{fig:figTranscribe4} and \ref{fig:figTranscribe5}).
The interface shows the transcription results to the users, and the users can optionally edit the transcription results to fix errors or make model improvements (Figure \ref{fig:figTranscribe6}).
The refined transcriptions can then be downloaded by the users.
If the user's data privacy preferences allow, the system can also collect them along with the original speech recordings as extra training data to further fine tune the model.

\subsection{Automatic Glossing}
\begin{figure*}[t]
\centering
\includegraphics[width=0.75\linewidth]{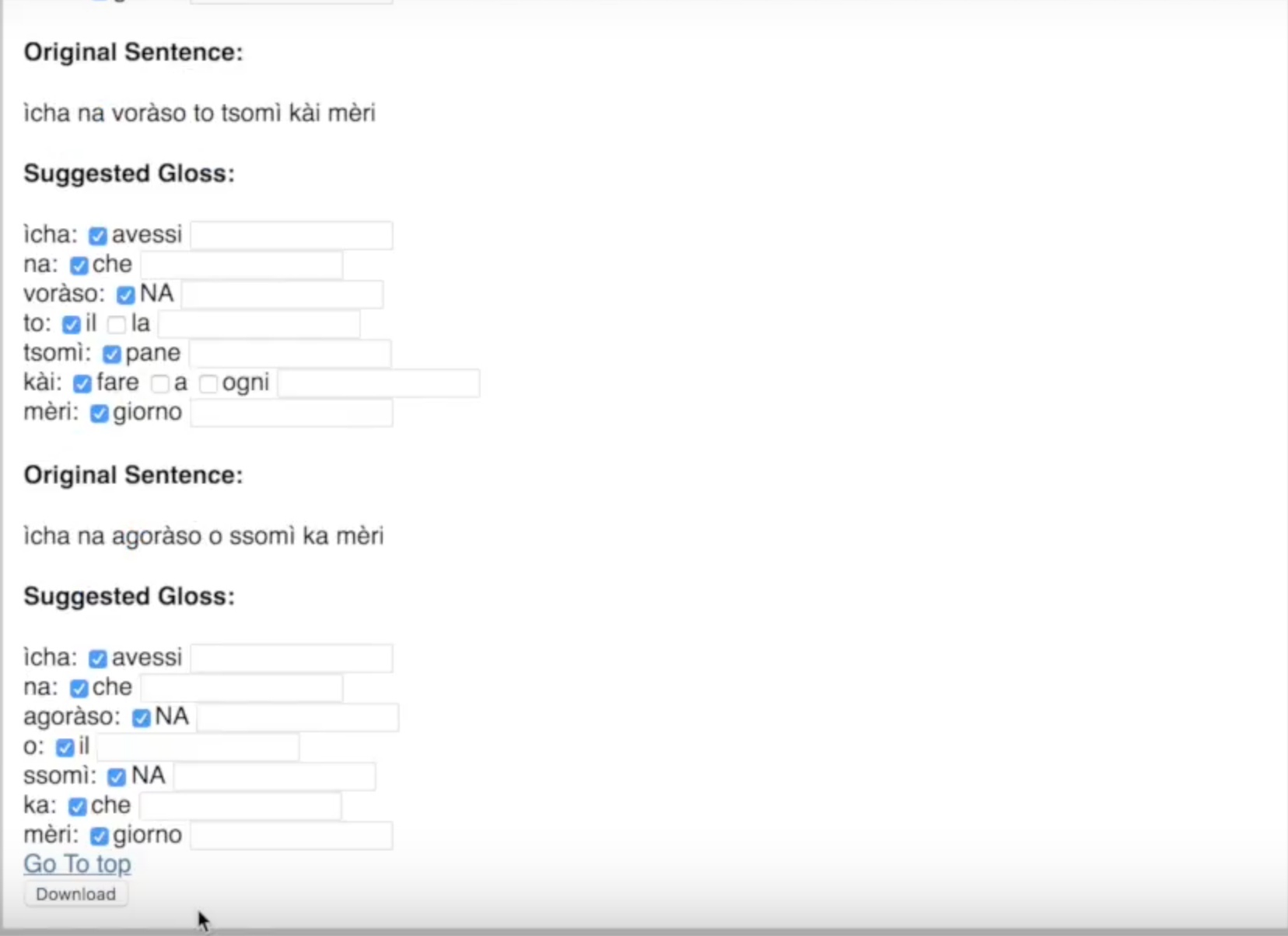}
\caption{An example of generated glosses, for Griko language data from \newcite{anastasopoulos2018griko}.}
\label{fig:figTranscribe7}
\end{figure*}

The interface also supports making glossing suggestions.
Glosses are generated word-by-word with Moses \cite{koehn07moses}, a statistical machine translation system.
The system takes parallel data as input, which could be either the  language and translations, or the language and glosses.
% The statistical model mainly takes the words and segments from the input sentences and infer the correspondences between the two languages through their co-occurrences. 
% The working flow of extracting the glossing suggestion dictionary by using Moses could be broken into the following steps: First input the parallel corpus, which is the translation between two languages.
Using this parallel data, we learn a word alignment with a statistical model, specifically the IBM alignment models \cite{brown93smt} as implemented in GIZA++ \cite{och03alignment}.
Then we perform phrase extraction \cite{koehn10smt}, which gives us a translation probability distribution for each word or phrase in the combined corpus.
We then display translations with high probability as glossing suggestions.
% TODO: example of glossing suggestions.
An example of how the automatic glossing suggestion works on the interface can be seen in Fig. \ref{fig:figTranscribe7}.
\section{Future Plans}
\label{sec:future}

Working with field data is highly rewarding, but on a moment-to-moment basis the work is not usually particularly \emph{engaging}; most of the individual decision events that a linguist makes during field corpus creation do not fully engage their reasoning capacity.
Our goal is to maximize the effects of human engagement with data by maximizing the time the linguist spends on interesting and relevant decisions.
We intend to explore this question with respect to both low-level decisions (``What word was said here?") and high-level decisions (``These utterances exemplify ergativity in this language; are there other examples in this corpus?").
Our future work towards this goal will take a three-pronged approach: developing a general-purpose linguistic annotation API and integrating it with popular annotation frameworks, developing new methods to perform multi-lingual and multi-task learning to train effective models even in a paucity of training data, and working with linguists to help refine and prioritize our work in these areas.
In particular, for the third goal we are actively seeking collaborators who would be interested in testing and giving advice about the utility of the proposed approach.

\section*{Acknowledgements}

We thank Alexis Michaud for his useful comments and help in preparation of data, Oliver Adams for his help with Persephone, and Antonis Anastasopoulos for helping us access and prepare the Griko data. 
This material is based upon work supported by the National Science Foundation under Grant No. 1761548.

\bibliography{myabbrv,gneubig}
\bibliographystyle{acl_natbib_nourl}

\end{document}